\documentclass{article}

\PassOptionsToPackage{round}{natbib}

\usepackage[main,final]{neurips_2026}

\usepackage[utf8]{inputenc}
\usepackage[T1]{fontenc}
\usepackage{hyperref}
\usepackage{url}
\usepackage{booktabs}
\usepackage{amsfonts}
\usepackage{nicefrac}
\usepackage{microtype}
\usepackage{xcolor}

\usepackage{amsmath}
\usepackage{amssymb}
\usepackage{mathtools}
\usepackage{amsthm}

\usepackage[capitalize,noabbrev]{cleveref}
\RemoveFromHook{label}[firstaid/cleveref]

\theoremstyle{plain}
\newtheorem{theorem}{Theorem}[section]

\newtheorem{lemma}[theorem]{Lemma}

\theoremstyle{definition}
\newtheorem{definition}[theorem]{Definition}

\theoremstyle{remark}
\newtheorem{remark}[theorem]{Remark}

\hypersetup{
    colorlinks,
    linkcolor=black,
    citecolor={blue!50!black},
    urlcolor={black}
}

\newcommand{\ve}[1]{\boldsymbol{#1}}

\newcommand{\R}{\mathbb{R}}

\renewcommand{\P}{\mathbf{P}}
\newcommand{\E}{\mathbf{E}} 
\newcommand{\1}{\mathbf{1}}

\newcommand{\grad}{\nabla}

\DeclareMathOperator{\Cov}{Cov}

\title{Susceptibilities for Neural Networks Learning from Physical Data}

\author{%
  Rohan Hitchcock\thanks{Work done while at the University of Melbourne and CSIRO.} \\
  Resolution \\
  \texttt{hitchcock@resolution.org} \\
  \And
  Gary W. Delaney\\
  CSIRO \\
  \texttt{gary.delaney@csiro.au} \\
  \And
  Jonathan H. Manton\\
  University of Melbourne\\
  \texttt{jmanton@unimelb.edu.au} \\
  \And
  Richard Scalzo\thanks{Work done while at CSIRO.} \\
  Monash University\\
  \texttt{richard.scalzo@monash.edu} \\
  \And
  Jingge Zhu\\
  University of Melbourne\\
  \texttt{jingge.zhu@unimelb.edu.au} \\
}

\begin{document}

\maketitle

\begin{abstract}
    Neural networks that are trained on data arising from a physical system must somehow learn regularities induced by the underlying physical laws. In this setting, concepts from statistical physics provide powerful tools for analysing how physical laws influence the learning process. In this paper we explain how \emph{susceptibilities} --- which measure the response of a neural network to perturbations of a parameter of the data distribution --- can be applied when learning from physical data. We show that such susceptibilities can identify the conditions under which specific input features are most informative for learning, guiding data selection, a prediction we validate experimentally. For distributions lacking a natural parameter, we propose introducing one via a physically meaningful scalar function of the physical state, and demonstrate that susceptibilities track the emergence of specific capabilities (such as collision detection) during training.
\end{abstract}

\section{Introduction} \label{sec:intro}

Distributions of physical data have rich and interesting structure arising from underlying 
physical laws. While many kinds of data exhibit a high degree of regularity, 
physical data is somewhat unique in the respect that the distribution is often parametrised 
by physically meaningful parameters. A typical situation is a Boltzmann distribution 
$q_h(x) \propto \exp(-\beta V(x; h))$ where $\beta > 0$ and the energy function $V(x; h)$ 
of a state $x$ depends on a parameter $h$. Statistical physics studies such distributions 
via expectations $\E_{x\sim q}[\psi(x)]$ and their derivatives with respect to parameters 
like $\beta$ and $h$. For example, the response of the Ising model (a model of magnetism) 
to perturbations in an external field controlled by $h$ is studied via the 
\emph{magnetic susceptibility} $\chi = \frac{\partial}{\partial h} \E_{x\sim q}[\psi(x)]$, 
where $\psi(x)$ is the magnetisation \citep[Chapter 2]{yeomansStatisticalMechanicsPhase1992}.

When learning from physical data, physical parameters in the data distribution are inherited 
by the learning process. For example, when learning to approximate the distribution $q_h(x)$
via a family of distributions $p(x;w)$ parametrised by $w\in \mathcal{W}$, the expected 
negative log-likelihood function 
\[
    L^h(w) = -\int _\mathcal{X} \log \bigl(p(x; w)\bigr) q_h(x)\, dx
\]
depends on $h$ via $q_h(x)$. It is natural to ask what role $h$ plays in the learning process. 

In this paper we study the Bayesian \emph{susceptibilities} framework, proposed recently by 
\citet{bakerStructuralInferenceInterpreting2025} in the context of language models, 
for neural networks trained on physical data. These measure the response of a neural 
network, as measured by a scalar function $\phi : \mathcal{W} \to \R$ of the model's 
parameters, to fluctuations in a physical parameter $h$ of the data distribution.

We show that susceptibilities can be used to identify conditions under which 
certain features of the input data are most informative for learning a 
particular task, in the setting of a transformer modelling the distribution of 
\emph{percolations} on a lattice. This suggests a way of selecting training data to improve 
learning; this claim is validated experimentally by retraining with differently 
sampled training data.

For distributions lacking an existing parameter, we propose an approach to introducing 
one via a physically meaningful scalar function of the physical state. In a transformer trained to 
simulate a granular material, we demonstrate that a sign change in the susceptibility coincides 
with the emergence of collision-resolution capability.

\section{Background} \label{sec:background}

\subsection{Singular learning theory} \label{sec:background-slt}
Singular learning theory (SLT) was developed as a theory of Bayesian learning 
\citep{watanabe2009GreyBook} and it is argued 
\citep[see][]{wei2022deepLearningSingular,lehalleurYouAreWhat2025} that SLT 
can be used as a theoretical framework for deep learning. 
Developments such as \emph{internal model selection} 
\citep{chenDynamicalBayesianPhase2023} and the \emph{local learning coefficient}
\citep{lau2024LLC} allow concepts from SLT to be applied to neural networks trained 
via stochastic optimization. 

Let $q(x)$ be a distribution on some set $\mathcal{X}$, 
which we aim to approximate. Let $\mathcal{W}$ be the parameter space of the model 
and consider distributions $p(x|w)$ parametrised by $w\in \mathcal{W}$ together 
with a prior distribution $\varphi(w)$ on $\mathcal{W}$. For example, $\mathcal{X}$
may be a set of sequences and $p(x|w)$ the probability assigned to a sequence 
$x\in \mathcal{X}$ by a transformer with parameters 
$w\in \mathcal{W}$. Let $\ell(x, w) = -\log p(x|w)$
and define the negative log-likelihood function $L(w) = \E_{x\sim q}[\ell(x, w)]$. 
Given an iid dataset $X_1, \ldots, X_n$
drawn from $q(x)$, define
$L_n(w) = \frac{1}{n}\sum _{i=1}^n \ell(X_i, w)$; the empirical version of $L(w)$. 
The posterior distribution is $p_n(w) \propto \exp(-n\beta L_n(w))\varphi(w)$.

Posterior probability mass concentrates in distinct regions called 
\emph{internal models}, and that the Bayesian learning process can be understood 
as selection between internal models as the dataset size $n$ is increased 
\citep{watanabe2009GreyBook,chenDynamicalBayesianPhase2023,lau2024LLC}.
\citet{chenDynamicalBayesianPhase2023,lau2024LLC} propose extending SLT to study 
neural networks trained via stochastic optimization where
local neighbourhoods of parameters along the 
training trajectory are treated as internal models of a Bayesian model. 
Throughout this paper we refer to the internal model associated to a parameter 
$w^*$ on the training trajectory as its \emph{local Bayesian model}
and study the behaviour of the network at $w^*$ via this object.

In practice, locality to $w^*$ is enforced using a Gaussian prior 
$\varphi(w)\propto \exp(-\tfrac{\gamma}{2}\|w - w^*\|^2)$,
where $\gamma > 0$ is a hyperparameter defining a soft localisation constraint,
and samples are collected from the posterior using a stochastic gradient Markov chain Monte 
Carlo algorithm. \citet{lau2024LLC,hitchcockGlobalLocalScalable2025} have 
validated that this approach is able to recover information about the local geometry 
of $w^*$, and other empirical work \citep{wang2025differentiation,hooglandLossLandscapeDegeneracy2025,urdshals2025structure,bakerStructuralInferenceInterpreting2025,gordon_towards_2026,wangEmbryologyLanguageModel2025}
has corroborated the utility of this perspective for 
understanding neural networks trained via stochastic optimization. 

\subsection{Susceptibilities} \label{sec:background-susceptibilites}

Susceptibilities are used to study how the model responds to 
variation in the data distribution.
Suppose that the data distribution $q_h(x)$ depends smoothly 
on a real parameter $h$. The negative log-likelihood function 
\[
    L^h(w) = \int \ell(x, w)q_h(x)\, dx
\]
also depends on $h$. Following \citet{bakerStructuralInferenceInterpreting2025}, 
for the purpose of defining susceptibility we consider the
\emph{quenched posterior} $p_h(w) \propto \exp(-n\beta L^h(w))\varphi(w)$
at inverse temperature $n\beta > 0$, defined in terms of $L^h(w)$
so that derivatives with respect to $h$ may be taken. Recall that 
$\varphi(w)$ is the Gaussian prior centred at some $w^* \in \mathcal{W}$. 

\begin{definition}[{\citealp{bakerStructuralInferenceInterpreting2025}}] \label{def:susceptibility}
    Let $\phi : \mathcal{W} \to \R$ be a function of the model weights. 
    The \emph{susceptibility of $\phi$ with respect to $h$} at inverse 
    temperature $n\beta > 0$ is 
    \begin{equation} \label{eqn:definition-of-susceptibility}
        \chi(h) = \frac{1}{n\beta} \frac{\partial}{\partial h} 
        \E _{w \sim p_h} \left[ \phi(w) \right] .
    \end{equation}
\end{definition}

The susceptibility measures the 
response in the (quenched) posterior to perturbations in $h$. 
In our experiments $\phi$ is always the loss on some dataset.
One can show that \citep[see][Lemma 2.2]{bakerStructuralInferenceInterpreting2025}
\begin{equation} \label{eqn:general-susceptibility-as-covariance}
    \chi(h) = -\Cov _{w\sim p_h} (\phi(w), \frac{\partial L^h}{\partial h}(w) ) + O(1/n) .
\end{equation}

In \citet{bakerStructuralInferenceInterpreting2025}, the family of distributions $q_h(x)$ 
is constructed such that setting $h=0$ recovers the training distribution $q_\text{train}(x)$ 
and $\chi(0)$ quantifies the response to a perturbation away from it. More generally, 
one may estimate $\chi(h)$ for $q_h\neq q_\text{train}$, as the data-refined local 
learning coefficient \citep{wang2025differentiation}; we use this generalisation 
in \cref{sec:percolation-theory}.
Other applications of susceptibilities are discussed and compared with our work in 
\cref{sec:related-work}.

\section{Susceptibilities in learning physical systems} \label{sec:susc-in-physical-systems}
We consider two settings: a physical system with a natural control parameter, demonstrated for 
percolations on a lattice (\cref{sec:percolation-theory}); and a physical system without one, 
where we propose introducing a control parameter via a scalar function of the state, demonstrated 
for granular materials (\cref{sec:granular-theory}).

\subsection{Percolation} \label{sec:percolation-theory}

We consider the distribution of \emph{percolations} on a $H\times H$ 
lattice with $N = H ^2$ sites. A state \emph{percolates} if there exists 
a connected component of occupied sites spanning from the bottom to the 
top of the lattice. A site is either 
\emph{occupied} or \emph{unoccupied}, and a \emph{state} is an assignment of 
`occupied' or `unoccupied' to each site. 
We represent a state by a vector $\ve x = (x_1, \ldots, x_N) \in 
\mathcal{X} = \{0, 1\}^N$ where $x_i = 1$ if site $i$ is occupied and $x_i=0$ if 
it is unoccupied. Let $K(\ve x) = \sum _{i=1} ^N x_i$ be the number of occupied sites
in a state $\ve x$. We define a distribution on $\mathcal{X}$ by
\begin{equation} \label{eqn:percolation-distribution-defn}
    q_h(\ve x) \propto \1(\ve x~\text{percolates}) h^{K(\ve x)} (1-h)^{N - K(\ve x)} . 
\end{equation}
That is, $q_h(\ve x)$ is obtained by conditioning the iid Bernoulli distribution 
where each site is occupied with probability $h$ on the event that the state $\ve x$
percolates. In experiments we sample from $q_h(\ve x)$ using rejection sampling 
(see \cref{apx:percolation-data-generation-and-critical-site} and \cref{fig:example-of-percolation}). For an introduction to percolation theory
see \citet{stauffer1985percolation}.

We regard $q_h(\ve x)$ as a distribution on binary strings of length 
$N$ by ordering the sites by first ordering the rows of the lattice from 
bottom to top and then ordering sites from left to right within 
each row. Let $p(\ve x | w)$ denote the distribution 
on $\mathcal{X}$ predicted by an autoregressive transformer with causal 
attention \citep[see][]{phuongFormalAlgorithmsTransformers2022}, 
where $w \in \mathcal{W} \subseteq \R^d$ are the trainable parameters. 
Let $L^h(w)$ and $p_h(w)$ be the expected negative log-likelihood and the 
local posterior distribution defined in \cref{sec:background} respectively, and 
$\chi(h)$ the susceptibility of a function $\phi : \mathcal{W} \to \R$ with 
respect to $h$.

\begin{lemma} \label{lemma:percolation-susceptibility-expression}
    The susceptibility $\chi(h)$ is given by
    \begin{equation} \label{eqn:percolation-susceptibility-nested-covariance}
        \chi(h) = \frac{-1}{h(1-h)}\Cov _{w\sim p_h} \bigl( \phi(w) , C_h(w) \bigr) + O(\tfrac{1}{n})
    \end{equation}
    where $C_h(w) = \Cov _{\ve x\sim q_{h}}\bigl(K(\ve x), \ell(\ve x, w)\bigr)$.
    \begin{proof}
        See \cref{sec:proofs}, \cref{lemma:restated-percolation-susceptibility-expression}.
    \end{proof}
\end{lemma}

\subsubsection{Masked susceptibilities} \label{sec:percolation-masked-susceptibilities}

Regarding states $\ve x$ as binary sequences using the ordering of sites described above,
let $p(x|x_1, \ldots, x_i; w)$ denote the probability the transformer predicts 
that site $i+1$ has occupation status $x\in \{0, 1\}$, so that 
\[
    p(\ve x|w) = p(x_1) \prod _{i=1} ^N p(x_{i+1}|x_1, \ldots, x_i; w) .
\]
We can break down contributions of physical features of a state $\ve x$ to the susceptibility $\chi(h)$. 
Let $M : \mathcal{X} \to 2^{\{1, \ldots, N\}}$ be a function from states to subsets of sites. 
We think of $M$ as defining a physical feature of a state $\ve x$; for example, $M(\ve x)$ 
might be the set of sites within the largest occupied cluster. Let
\[
    \ell(\ve x, w; M) = -\sum _{i\in M(\ve x)} \log p(x_{i+1}|x_1, \ldots, x_i; w) 
\]
denote the contribution to the loss from sites within the mask $M$, and
$K(\ve x; M) = \sum _{i \in M(\ve x)} x_i$
denote the number of occupied sites within $M$. 
\begin{definition} \label{def:percolation-masked-susceptibility}
    Given $M, M' : \mathcal{X} \to 2 ^{\{1, \ldots, N\}}$, the \emph{masked susceptibility}
    $\chi(h; M, M')$ is defined as
    \[
        \chi(h; M, M') = \frac{-1}{h(1-h)}\Cov _{w\sim p_h} \bigl( \phi(w) , C_h(w; M, M') \bigr)
    \]
    where $C_h(w; M, M') = \Cov _{\ve x\sim q_{h}} \bigl(K(\ve x; M), \ell(\ve x, w; M')\bigr)$.
\end{definition}
We use masked susceptibilities to understand the contribution to $\chi(h)$ from different physical features of the input 
state. Given $M$, define $M^c(\ve x) = \mathcal{X} \setminus M(\ve x)$. From 
\cref{lemma:percolation-susceptibility-expression}, using the bilinearity 
of covariance and the decompositions $K(\ve x) = K(\ve x; M) + K(\ve x; M^c)$ 
and $\ell(\ve x, w) = \ell(\ve x, w; M) + \ell(\ve x, w; M^c)$, we obtain
\[
    \chi(h) = \chi(h; M, M) + \chi(h; M, M^c) + \chi(h; M^c, M) + \chi(h; M^c, M^c) .
\]

\subsection{Granular materials} \label{sec:granular-theory}

\subsubsection{Introducing a control parameter} \label{sec:granular-theory-introducing-control-parameter}
We begin by considering the general setting of
a data distribution $q(x)$ on a space $\mathcal{X}$, understood as 
a distribution on the set of states of some physical system. Let
$\psi: \mathcal{X} \to \R$ be a function which satisfies
\begin{equation} \label{eqn:exponential-moment}
    \int _\mathcal{X} q(x) e^{h\psi(x)}\, dx < \infty
\end{equation}
for all $h$ in some open neighbourhood of zero. We can introduce a control parameter $h$ by 
considering the distribution 
\begin{equation} \label{eqn:exponential-tilt-distribution}
    q_h(x) \propto q(x) e^{h\psi(x)} .
\end{equation}
Setting $h=0$ recovers the original distribution $q(x)$. 
The condition \eqref{eqn:exponential-moment} is a technical requirement which ensures 
that the susceptibility below is well defined, but in practice $\psi$ should additionally be 
chosen to be a physically meaningful observable of the system, so that perturbations in $h$ 
are physically meaningful. 
We discuss the choice of $\psi$ in practice in 
\cref{sec:choosing-observables}. Given a function 
$\phi : \mathcal{W} \to \R$ of the model parameters we consider susceptibility $\chi(h)$ of 
$\phi$ with respect to $h$, evaluated at $h=0$. Let $\chi = \chi(0)$. 

\begin{lemma} \label{lemma:psi-susceptibility-expression}
    We have
    \begin{equation} \label{eqn:psi-susceptibility-nested-covariance}
        \chi = -\Cov _{w\sim p} \bigl( \phi(w) , C(w) \bigr) + O(\tfrac{1}{n})
    \end{equation}
    where $p$ is the quenched posterior distribution on $\mathcal{W}$ at $h=0$ and
    $C(w) = \Cov _{ x\sim q}\bigl(\psi( x), \ell( x, w)\bigr)$.
    \begin{proof}
        See \cref{sec:proofs}, \cref{lemma:restated-psi-susceptibility-expression}.
    \end{proof}
\end{lemma}
Note that the expression \eqref{eqn:psi-susceptibility-nested-covariance} for $\chi$ 
has the same nested covariance structure as the expression 
\eqref{eqn:percolation-susceptibility-nested-covariance} for the susceptibility in the 
percolation model case. 

\subsubsection{Granular material physics}

We consider a system of two-dimensional identical spherical particles of radius $r$, 
confined to a rectangular box with constant gravitational acceleration. Collisions between 
particles are governed by \emph{damped linear spring dynamics} 
\citep[see][Section 2.3.1]{luding2008introduction}; 
the force experienced by a pair of colliding particles is computed as 
damped linear spring compressed to a depth
\begin{equation} \label{eqn:particle-overlap-definition}
    \delta _{ij}(x) = \max(0, 2r - d_{ij}(x))
\end{equation}
where $d_{ij}(x)$ is the distance between particle $i$ and particle $j$.
Note that two particles apply a force on one another if they overlap in space.
This can be interpreted as a simple 
model for deformation at the contact point for real-world collisions.
Collisions with boundaries are handled in the same way except that the boundary 
has ``infinite mass''; it does not move when hit by a particle. 

The state of the granular material consists of the position and momentum of 
each particle, and so $\mathcal{X}$ can be understood as a 
subset of $\R^{4N}$ where $N$ is the number of particles. A distribution 
$q(x)$ on $\mathcal{X}$ is induced by generating initial states according to some chosen 
distribution and allowing the system to evolve according to damped linear spring 
dynamics for some fixed duration. We show examples of granular material states in 
\cref{fig:example-of-granular-material}.

We train a neural network to predict the future state after some fixed interval 
$\Delta t$, given the current state of the system. This is a regression 
task and we train the model using mean-squared error loss. To apply susceptibilities, 
we formulate this as a Bayesian learning task by defining 
\[
    p(x|w) \propto \exp\bigl( -\|G(x) - F(x; w)\|^2 \bigr)q(x)
\]
where $F(x;w)$ denotes the function defined by the neural network 
with parameters $w \in\mathcal{W}$ and $G(x)$ denotes the result 
of allowing $x$ to evolve with damped linear spring dynamics for 
a duration $\Delta t$; i.e. $G(x)$ is the `true' function the 
network is being trained to learn. 
In this case, the quenched posterior is 
$p(w) \propto \exp(-n\beta L(w)) \varphi(w)$
where $L(w)$ is the expected mean-squared error on $q(x)$. 

\begin{figure}[t]
  \centering
  \includegraphics[width=\linewidth]{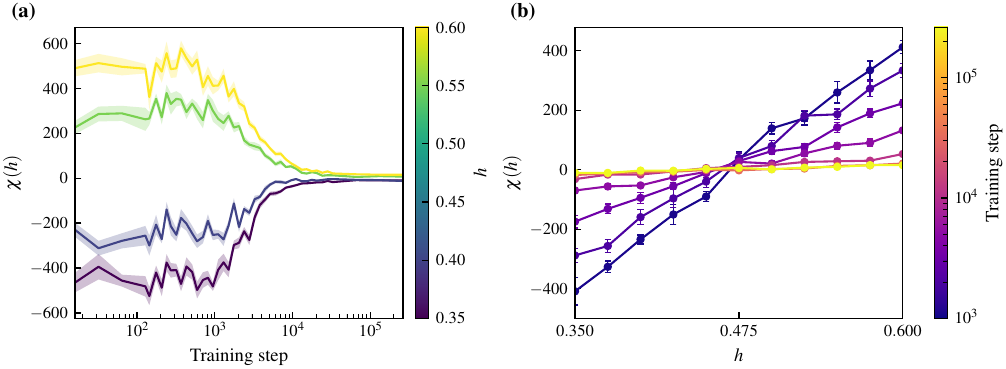}
  \caption{
        The susceptibility $\chi(h)$ shows how the percolation model
        described in \cref{sec:percolation-experiments} becomes calibrated 
        over all values of $h$ in the training distribution 
        \eqref{eqn:percolation-training-distribution}; contrast this with 
        $\chi_\text{crit.}(h)$ in \cref{fig:susc-critical-percolation} which
        shows the difficulty of learning percolation at different values of $h$.
        \textbf{(a)} $\chi(h)$ over training for various values of 
        $h$. \textbf{(b)} $\chi(h)$ plotted as a function of $h$ at 
        various points over training. Observe that $\chi(h)$ is roughly symmetric about 
        $h\approx 0.475$ --- mean value of $h$ in the training distribution ---
        at each point in training, and that $\chi(h)$ decays to zero 
        over training for all $h \in [0.35, 0.6]$.
  }
  \label{fig:susc-full-percolation}
\end{figure}

\section{Experiments} \label{sec:experiments}

\subsection{Percolation experiments} \label{sec:percolation-experiments}

Consider the setting of percolations described in \cref{sec:percolation-theory}. 
We train a 15M parameter transformer model on a mixture 
of percolating distributions $q_h(\ve x)$ on a $16\times 16$ lattice.
Specifically, 
the training distribution is 
\begin{equation} \label{eqn:percolation-training-distribution}
    q(\ve x) = \int _{0.35} ^{0.6} q_h(\ve x) u(h)\, dh
\end{equation}
where $u(h)$ is a probability density supported on $[0.35, 0.6]$. In all except for a 
few specific experiments we take $u(h)$ to be the uniform distribution. 
We use a standard transformer architecture; details are 
given in \cref{apx:percolation-training-and-model-details}. 
We treat each state $\ve x$ as a sequence using the ordering described in 
\cref{sec:percolation-theory}; rows are ordered from bottom to top and 
then within each row sites are ordered from left to right. This 
ordering determines which sites are `visible' when predicting a given 
site. 

\subsubsection{Estimating susceptibilities}

To understand how the model responds to different parts of the training 
distribution, we estimate the susceptibility $\chi(h)$ of $\phi(w) = L^h(w)$
with respect to the parameter $h$ in $q_h(\ve x)$ for different values 
of $h\in [0.35, 0.6]$. Since we want to understand how the model responds 
to different data at different points in training, we estimate 
the susceptibility at various training checkpoints. 

In addition to $\chi(h)$, we also estimate a masked susceptibility $\chi_\text{crit.}(h)$ 
which
is designed to isolate the effect of features related to percolation. 
This is motivated by the observation that, 
conditioning on sites within the transformer's causal mask, the majority of 
sites are unconstrained in the sense that their occupation status 
is not relevant to whether the state percolates, and predictions at these 
sites dominate the loss at all points in training. 
Sites involved in percolation are constrained to varying degrees, and
certain sites are so constrained that they must be occupied in order for the 
state to percolate. 

\begin{definition} \label{def:critical-site}
    Let $\ve x \in \mathcal{X}$ be a percolating state. A \emph{critical site} is a site 
    which, given the occupation status of sites in the model's causal mask, must be 
    occupied for $\ve x$ to percolate. 
\end{definition}

We define
$\chi_\text{crit.}(h) = \chi(h; \1, M_\text{crit.})$ 
(see \cref{def:percolation-masked-susceptibility}) where 
$M_\text{crit.}(\ve x)$ is the set of all critical sites in a percolating state $\ve x$
and $1(\ve x)$ denotes the set of all sites. Recall that $M_\text{crit.}$ selects the 
sites included in the sum of per-site losses, so 
$\chi_\text{crit.}(h)$ can be understood as the contribution 
to $\chi(h)$ from the per-site losses at critical sites. 

We estimate $\chi(h)$ and $\chi_\text{crit.}(h)$ using the
expression derived in \cref{lemma:percolation-susceptibility-expression}. 
The covariance over the posterior distribution $p_h(w)$ is estimated 
via stochastic gradient MCMC; see 
\cref{apx:percolation-estimating-susceptibilities-details} for details. 

\begin{figure}[t]
  \centering
  \includegraphics[width=\linewidth]{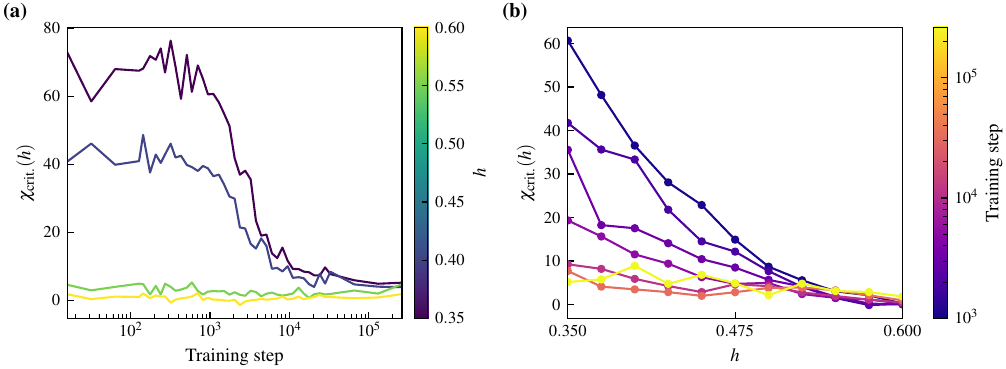}
  \caption{
        The masked susceptibility $\chi_\text{crit.}(h)$ (see \cref{sec:percolation-experiments})
        shows the difficulty of learning percolation at different values of $h$; contrast 
        this with $\chi(h)$ in \cref{fig:susc-full-percolation}.
        \textbf{(a)} $\chi_\text{crit.}(h)$ over training for various values of 
        $h$. \textbf{(b)} $\chi_\text{crit.}(h)$ plotted as a function of $h$ at 
        various points over training. Unlike $\chi(h)$, 
        $\chi_\text{crit.}(h)$ is non-negative for all 
        values of $h \in [0.35, 0.6]$ and is a decreasing function of $h$ at 
        each point in training. 
        We interpret this as indicating that smaller values of $h$
        make it easier to decrease the loss via improving prediction about 
        percolation. This inference is supported by retraining the model on 
        datasets biased toward smaller or larger values of $h$; see 
        \cref{fig:percolation-data-biasing-results}.
  }
  \label{fig:susc-critical-percolation}
\end{figure}

\subsubsection{Results}

The key finding is that $\chi(h)$ and $\chi_\text{crit.}(h)$ tell different stories:
the former reflects calibration across the training mixture and the latter reflects 
the difficulty of learning percolation at different $h$.
In \cref{fig:susc-full-percolation} we see that $\chi(h)$
is negative for $h < h_\text{mid}$, positive for $h > h_\text{mid}$ and 
roughly symmetric about $h_\text{mid}$ over training, where $h_\text{mid} \approx 0.475$. 
Over training $\chi(h)$ appears to decay to zero for all values of $h$. 
On the other hand, $\chi_\text{crit.}(h)$ is positive over training for all values 
of $h$ and appears to decay to zero. These results are displayed in \cref{fig:susc-critical-percolation}.

We interpret \cref{fig:susc-full-percolation} as the model becoming calibrated across the 
training distribution. The expected value of $L^h(w)$ over the local Bayesian model measures how 
well the model fits $q_h(\ve x)$; smaller is better. $\chi(h) > 0$ at $h > h_\text{mid}$ and 
$\chi(h) < 0$ at $h < h_\text{mid}$ early in training reflects that the model is best adapted 
near $h \approx 0.475$ --- the mean of the training mixture --- and the decay of $\chi(h)$ 
toward zero reflects calibration across the full range. This behaviour reflects the mixture 
training distribution and is not specific to percolation.

In contrast, $\chi_\text{crit.}(h) > 0$ across all $h$ indicates that critical sites contribute 
more to decreasing the posterior expectation of $L^h(w)$ as $h$ decreases. We interpret this as: 
smaller $h$ makes it easier for the model to improve via better prediction about percolation. 
Note that the model must both predict unconstrained sites at the correct Bernoulli probability 
$h$ and ensure that overall configurations percolate (these tasks are not independent).

\begin{figure}[t]
  \centering
  \includegraphics[width=3.25in]{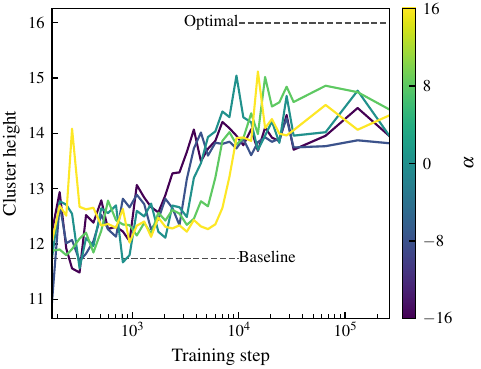}
  \caption{
        Biasing the training data towards smaller values of $h$ makes learning percolation 
        easier. We retrain the model with different training datasets parametrised by 
        $\alpha$ and track the mean maximum cluster height generated by the 
        model as a downstream measure of to what degree the model has learned percolation.
        When the dataset is biased toward small $h$ ($\alpha < 0$) the model learns 
        percolation faster than when the dataset is biased towards large $h$ ($\alpha > 0$). 
        For reference, we show the mean maximum cluster height when the model is replaced 
        with a random Bernoulli model (baseline) and the optimal value for a model which 
        always generates percolating configurations (optimal). 
  }
  \label{fig:percolation-data-biasing-results}
\end{figure}

To validate the prediction that smaller $h$ is easier for percolation, we 
retrain the model on biased distributions of the form 
\eqref{eqn:percolation-training-distribution} 
with $u(h) \propto \exp(\alpha h)$ 
on $[0.35, 0.6]$ ($\alpha < 0$ weights toward small $h$, $\alpha > 0$ toward 
large $h$, and $\alpha \to 0$ recovers the uniform mixture). We track the 
model's ability to generate percolating states via the mean 
\emph{maximum cluster height} of generated sequences 
(see \cref{apx:percolation-evaluating-generated-cluster-height}). 
\cref{fig:percolation-data-biasing-results} shows that models with $\alpha < 0$ 
learn percolation faster, and $\alpha > 0$ learn slower — consistent with our interpretation
of $\chi_\text{crit.}$.

We also compared $\chi_\text{crit.}(h)$ to other diagnostic values:
the empirical loss $L^h_n$ on iid samples drawn from $q_h$, the norm $\|\grad L_n^h\|$ of the gradient of this loss, 
and the dot product $\grad L_n ^h \cdot \grad L_\text{crit.}$ between the gradient of $L_n^h$ and the 
gradient of the loss $L_\text{crit.}$ at critical sites in samples drawn from the training distribution $q$. 
These do not predict the same data selection result as cleanly as $\chi_\text{crit.}(h)$, though 
$\grad L_n ^h \cdot \grad L_\text{crit.}$ does have some similar structure. We discuss this 
in more detail in \cref{apx:compare-percolation-baselines}.

\subsection{Granular material experiments} \label{sec:granular-material-experiments}

Consider the setting of \cref{sec:granular-theory}. We train a transformer model to 
predict the future state of a granular material with damped linear spring dynamics 
after a short duration $\Delta t$, given an input state $x$. The data distribution $q(x)$ 
is defined by randomly generating a particle state (with non-overlapping particles) and simulating 
damped linear spring dynamics, taking snapshots of particle states at regular intervals. The parameters 
of the particle simulation are given in \cref{apx:granular-data-generation}.

In the transformer, each  particle is processed in its own residual stream and attention 
heads are used to transfer information between particles. We use a variant of attention 
where the sum of the attention scores can be less than one, but otherwise the transformer 
architecture and training setup is standard; see \cref{apx:granular-training-and-model-details} 
for details.

\subsubsection{Susceptibilities for granular materials}

In the granular materials setting there is no natural control parameter, so we 
demonstrate that one can be introduced via a physically meaningful scalar 
function $\psi(x)$ as described in \cref{sec:granular-theory-introducing-control-parameter}.
Because the main difficulty in predicting future states of a granular material is 
resolving collisions between two or more particles, we take
$\psi(x) = \sum _{i=1} ^N \sum _{j=1} ^N f_{ij} (x)$
where $f_{ij}(x) \geq 0$ is the magnitude of the force exerted on particle $j$ by 
particle $i$. This is computed as $f_{ij}(x) = k\delta_{ij}(x)$ where $\delta_{ij}(x)$ 
is the overlap between particles $i$ and $j$ given in 
\eqref{eqn:particle-overlap-definition} and $k$ is the collision spring constant. 

Attention heads are expected to be essential for enabling the collision resolution 
capability of the model, given the model architecture. If susceptibilities defined 
using $\psi(x)$
are informative, they should detect when this capability emerges. 
Following \citet[Definition 2.4]{bakerStructuralInferenceInterpreting2025}, 
we consider \emph{component-wise loss} observable $\phi_C(w) = \delta_C(w)L(w)$, where $C$ 
is an architectural component of the model and $\delta_C(w)$ is a delta function-like 
term which has the effect of restricting posterior expectation integrals 
to only integrate over parameters in $C$; details are given in 
\cref{apx:granular-susceptibilities-method}. 
In our experiments with granular models we take $C$ to be the set of all 
attention heads in the model. This allows us to study how the loss varies over the 
posterior due to changes exclusively within attention head parameters. The corresponding 
susceptibility quantifies how this varies in response to perturbations in $h$.

Let $\chi_{\text{heads}}$ denote the susceptibility of $\phi_C(w)$ with respect to the control parameter 
$h$ in \eqref{eqn:exponential-tilt-distribution} at $h=0$, where $\psi(x)$ 
is the total collision force magnitude above. Similarly, let $\chi_\text{full}$ denote the 
susceptibility of $\phi(w) = L(w)$; that is, the susceptibility without the restriction to attention 
heads. We can estimate $\chi_{\text{heads}}$ and $\chi_{\text{full}}$ using the nested covariance expression 
\eqref{eqn:psi-susceptibility-nested-covariance} derived in 
\cref{lemma:psi-susceptibility-expression}. 

\subsubsection{Results}

\begin{figure}[t]
  \centering
  \includegraphics[width=3.25in]{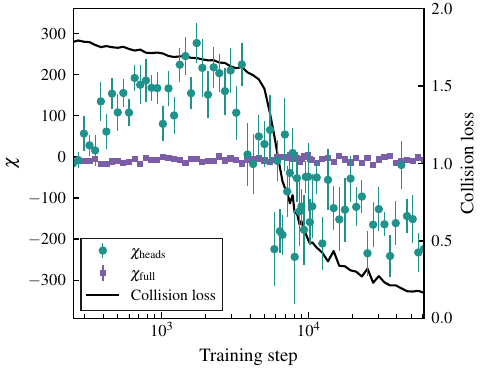}
  \caption{
        The change in sign of the susceptibility $\chi_{\text{heads}}$ over training reflects the development of
        collision resolution capability by the model. Early in training $\chi_{\text{heads}}$ (the 
        model's susceptibility to the total collision force) is positive, reflecting that the 
        model is less suitable for states with more collisions. Later in training it 
        becomes negative at the same time that the collision loss (the mean-squared error 
        of predicted vs. actual states of colliding particles) decreases rapidly. The full susceptibility 
        $\chi_{\text{full}}$ does not show this signal. 
  }
  \label{fig:susc-granular}
\end{figure}

We see in \cref{fig:susc-granular} that $\chi_{\text{heads}}$ shows that at a certain point in training, 
the model changes from being poorly suited to states with many collisions to being 
well-suited to such states. This coincides with the development of collision resolution 
capabilities as measured by the loss on colliding particles. Additionally, since 
we use the component-wise loss restricted to attention heads, this confirms attention 
heads contribute to this capability developing. 

More specifically, the susceptibility is positive early in training, indicating that decreasing 
$h$ below zero (i.e. weighting the data away from states where $\psi(x)$ is large) decreases the 
expected component-wise loss. We interpret this as indicating that the attention heads have not 
yet learned how to correctly resolve collisions between particles and so the model is better 
suited to a dataset with fewer collisions. Later in training, $\chi_{\text{heads}}$ changes quickly to a 
negative value, indicating that increasing $h$ (i.e. weighting the data towards states where 
$\psi(x)$ is large) decreases the expected component-wise loss. In contrast, $\chi_\text{full}$ 
stays flat over training and does not undergo a clear sign change that coincides with the 
emergence of collision detection behaviour. 

\section{Discussion} \label{sec:discussion}

In percolation models, we showed that susceptibilities are able to identify 
components of the training distribution that are most informative for learning 
percolation. While the overall susceptibility reflects calibration over the range of parameter 
values in the training distribution, the contribution to the susceptibility from 
critical sites suggests that smaller parameter values make it easier to 
learn features related to percolation. 
This demonstrates how susceptibilities can be used to inform training 
data selection so as to promote or inhibit the model developing a given capability;
for percolation models, the ability to generate states which percolate. 
Retraining on datasets biased toward (resp. away) from smaller parameter values 
accelerates (resp. decelerates) learning percolation capabilities, supporting 
our interpretation of susceptibilities at critical sites. 

We further propose a general approach to defining susceptibilities via a scalar function 
$\psi(x)$ of the data (\cref{sec:granular-theory-introducing-control-parameter}). 
The construction applies to any data distribution and any $\psi$ satisfying the technical 
condition \eqref{eqn:exponential-moment} --- though $\psi$ should be chosen to be a 
physically meaningful observable (\cref{sec:choosing-observables}) --- and the resulting susceptibility admits the same 
nested-covariance form as in the percolation case (\cref{lemma:psi-susceptibility-expression}). We 
demonstrate the construction in a transformer trained to simulate a granular material, taking 
$\psi(x)$ to be the total magnitude of all collision forces in a state. The susceptibility 
undergoes a sign change that coincides with the emergence of collision-resolution capability.

\subsection{Choosing observables in practice} \label{sec:choosing-observables}

In choosing an observable $\psi : \mathcal{X} \to \R$ to define a susceptibility, we have 
in mind a practitioner who knows some property of the physical system they want the model 
to represent, and designs $\psi$ to capture it. This choice does not need to be made once, 
or correctly on the first attempt: most of the cost of estimating the susceptibility comes from 
posterior sampling, and choosing several observables $\psi_1, \ldots, \psi_P$ only requires
computing additional data covariances $\Cov_{x\sim q}(\psi_i(x), \ell(x, w))$ at each sampling 
step. There is little additional cost to choosing many observables and analysing the resulting 
vector of susceptibilities.

Candidate observables can often be generated without deep domain knowledge. Several observables 
can serve as a proxy for the same behaviour: for example, while for the granular system we used total 
collision force, but quantities such as the mean overlap and mean number of particle contacts all 
capture interactions between particles. Another reasonable candidate are quantities which govern the 
crossover from one regime to another, such as the Reynolds number in fluid mechanics. Likewise, quantities 
related to the modelling application such as the stress or strain on a solid body due to mechanical or thermal loads.

\subsection{Related work} \label{sec:related-work}

Susceptibilities in the general sense were introduced 
by \citet{bakerStructuralInferenceInterpreting2025}, and prior work has applied 
the concept to different kinds of perturbations of the data distribution.
In \citet{bakerStructuralInferenceInterpreting2025,wangEmbryologyLanguageModel2025,gordon_towards_2026}
a parameter $h$ controls the mixture of two datasets $q_h(x) = (1-h)q(x) + hq'(x)$
in the context of language modelling. These are used to study the response of 
components of the model to different types of data (e.g. code vs. prose). 
In 
\citet{kreerBayesianInfluenceFunctions2025,leeInfluenceDynamicsStagewise2025,adamLossKernelGeometric2025} 
the perturbation parameter $h$ is the importance of a particular sample $x$
and $\phi(w)$ the loss of another sample $x'$. This instantiation of susceptibilities 
is referred to as a \emph{local Bayesian influence function}, and is proposed as a 
method for data attribution. In contrast to this prior work (which considers data-mixture 
and sample-importance perturbations) we study susceptibilities derived from 
physically meaningful perturbations of the data distribution, in the context of 
learning from physical data. These arise via either an existing control parameter
or a function $\psi$ of the physical system; the latter can be applied to any data 
distribution.

Analogies between statistical physics and statistical inference have been 
noted by many \citep[see for example][]{zdeborova2016statistical,lamont2019correspondence}.
We exploit a direct connection when the training data itself comes 
from a physical system. In this context, we explain how perturbations of physical parameters 
--- which induce a response in the physical system --- similarly 
induce a response in the model, and that this connection can be used to make decisions about 
training.

\subsection{Limitations} \label{sec:limitations}

The sampling protocol which is used to estimate susceptibilities has hyperparameters
(sampler step size, $\gamma$, and $n\beta$) which must be jointly tuned to ensure stable sampling. 
This requirement substantially constrains the available choices.
We did not conduct a systematic study of robustness to sampler hyperparameters. Our procedure 
for selecting them uses diagnostics of sampler health that are independent of the 
susceptibility (see \cref{apx:sampler-tuning}); a systematic sensitivity analysis is an 
important direction for future work. The measurements 
we report are corroborated by their predictive validity: in \cref{sec:percolation-experiments}
the susceptibility correctly predicts which training distributions accelerate learning and in 
\cref{sec:granular-material-experiments} the susceptibility tracks 
the independently measurable emergence of collision resolution.

The experimental evaluation in this paper is limited to two synthetic physical systems. 
Broader empirical validation in more complex or naturalistic settings is a natural direction for future work.

\subsection{Future directions}

It is natural to consider the effect of phase transitions and 
critical phenomena in the data distribution on the learning process, and 
in particular upon susceptibilities. 
The percolation system described in \cref{sec:percolation-theory}, 
for example, has a phase transition at $h = h_c\approx 0.59275$ 
(\citealp[Table 1]{stauffer1985percolation}; see \cref{remark:on-critical-threshold}). 
We did not see any special behaviour near $h_c$ (we did not specifically design
experiments to investigate critical phenomena), likely due to finite size 
effects; critical phenomena only truly occur as $N \to \infty$ where $N$ 
is the size of the lattice. In our percolation experiments, $N$ is relatively small 
and scaling $N$ means scaling the context window of the model, which introduces additional cost 
and complexity.

The $\psi$-based construction offers an 
alternative to the data mixture approach described in 
\citet{bakerStructuralInferenceInterpreting2025}.
Our approach may offer advantages when it is difficult to
curate a dataset to target a certain 
property. Whether these approaches offer different trade-offs or are complementary 
remains to be investigated.

\begin{ack}
RH was supported by CSIRO through the AI for Missions PhD Program and by the Commonwealth 
through an Australian Government Research Training Program Scholarship 
(DOI: \url{https://doi.org/10.82133/C42F-K220}). Computing resources were provided by CSIRO. 
\end{ack}

\clearpage 

\bibliographystyle{plainnat}
\bibliography{main}

\clearpage
\appendix

\section*{Appendix}
These appendices contain proofs and additional methodological details, supporting the main text.
\begin{itemize}
    \item \cref{sec:proofs} gives proofs for the theoretical results stated in \cref{sec:susc-in-physical-systems}.
    \item \cref{apx:method-details} provides additional details about the methodology of the 
    experiments discussed in \cref{sec:experiments}.
    \item \cref{apx:sampler-tuning} gives detail on sampling and hyperparameter selection. 
\end{itemize}

\clearpage

\section{Proofs} \label{sec:proofs}

\begin{lemma}[{restating \cref{lemma:percolation-susceptibility-expression}}] 
\label{lemma:restated-percolation-susceptibility-expression}
    Consider the setting of \cref{sec:percolation-theory} and let $\phi : \mathcal{W} \to \R$ be 
    a function of the model's weights. The 
    susceptibility $\chi(h)$ of $\phi$ with respect to $h$ (see 
    \cref{def:susceptibility})
    is given by
    \[
        \chi(h) = \frac{-1}{h(1-h)}\Cov _{w\sim p_h} ( \phi(w) , \Cov _{\ve x\sim q_{\ve h}}(K(\ve x), \ell(\ve x, w))) + O(\tfrac{1}{n}) .
    \]
    where $K(\ve x) = \sum _{i=1} ^N x_i$ is the number of occupied sites in a given state. 
    \begin{proof}
        Using \eqref{eqn:general-susceptibility-as-covariance} we have
        \[
            \chi(h) = -\Cov _{w\sim p_h} (\phi(w), \frac{\partial L ^h}{\partial h}(w)) + O(\tfrac{1}{n}).
        \]
        We have 
        \begin{align*}
            \frac{\partial L^h}{\partial h}(w) = \sum  _{\ve x \in \mathcal{X}} \ell(\ve x, w) \frac{\partial}{\partial h} q_h(\ve x) &= - \sum  _{\ve x \in \mathcal{X}} \log p(\ve x|w) \frac{\partial}{\partial h} q_h(\ve x). 
        \end{align*}
        Let $A\subseteq \mathcal{X}$ be the subset of all percolating configurations, 
        and $K(\ve x) = \sum _{i=1} ^N x_i$ the number of occupied sites in a state 
        $\ve x \in\mathcal{X}$. We have 
        \begin{align*}
            q_h(\ve x) &= \1(\ve x \in A) \frac{h^{K(\ve x)}(1 - h) ^{N-K(\ve x)}}{\P _h(A)}
        \end{align*}
        where we define $\P _h(A) = \sum _{\ve x \in A} h^{K(x)} (1-h) ^{N-K(x)}$. Taking 
        derivatives with respect to $h$, we have:
        \begin{align*}
            \frac{\partial }{\partial h} \left(\1(\ve x\in A) h ^{K(\ve x)} (1-h) ^{N-K(\ve x)}\right) 
            &= \1(\ve x\in A) \big( K(\ve x) h^{K(\ve x) - 1} (1-h) ^{N-K(\ve x)} \\
            &\qquad - (N-K(\ve x)) h^{K(\ve x)} (1-h) ^{N-K(\ve x) - 1} \big)\\
            &= \frac{1}{h} K(\ve x) \P _h (A) q_h(\ve x) - \frac{1}{1-h} (N-K(\ve x)) \P_h(A) q_h(\ve x) \\
            &= \P_h(A) q_h(\ve x) \left(\frac{K(\ve x)}{h(1-h)} - \frac{N}{1 - h}\right) 
        \end{align*}
        and 
        \begin{align*}
            \frac{\partial }{\partial h} \P _h(A) &= \sum _{\ve x\in A} \frac{\partial }{\partial h} \left(h^{K(\ve x)} (1-h) ^{N-K(\ve x)} \right) \\
            &= \sum _{\ve x\in A} K(\ve x) h ^{K(\ve x)-1}(1-h) ^{N-K(\ve x)} -  (N - K(\ve x)) h^{K(\ve x)} (1-h) ^{N-K(\ve x) - 1} \\
            &= \sum _{\ve x\in A} \frac{1}{h} K(\ve x) \P_h(A) q_h(\ve x) -  \sum _{\ve x\in A}  \frac{1}{1-h}(N - K(\ve x)) \P_h(A) q_h(\ve x) \\
            &= \frac{1}{h} \E_h[K(\ve x)] \P_h(A) -  \frac{1}{1-h}\E_h[N - K(\ve x)] \P_h(A) \\
            &= \P_h(A) \left(\frac{\E_h[K(\ve x)]}{h(1-h)} - \frac{N}{1-h}\right) .
        \end{align*}
        It follows from the product rule that 
        \begin{align*}
            \frac{\partial}{\partial h}q_h(\ve x) &= \frac{1}{\P_h(A) ^2} \bigg(
                \P_h(A)^2 q_h(\ve x) \left(\frac{K(\ve x)}{h(1-h)} - \frac{N}{1 - h}\right) \\
                &\qquad - \1(\ve x \in A) h^{K(\ve x)} (1-h) ^{N-K(\ve x)} \P_h(A) \left(\frac{\E_h[K(\ve x)]}{h(1-h)} - \frac{N}{1-h}\right)
                \bigg) \\
                &= q_h(\ve x) \left(\frac{K(\ve x)}{h(1-h)} - \frac{N}{1 - h}\right) - q_h(\ve x) \left(\frac{\E_h[K(\ve x)]}{h(1-h)} - \frac{N}{1-h}\right) \\
                &= \frac{q_h(\ve x)}{h(1-h)} \left(K(\ve x) - \E_h[K(\ve x)]\right) .
        \end{align*}
        Hence we have 
        \begin{align*}
             \frac{\partial L^h}{\partial h}(w) &= \sum  _{\ve x \in \mathcal{X}} \ell(\ve x, w) \frac{\partial}{\partial h} q_h(\ve x) \\
             &= \frac{1}{h(1-h)} \Cov _{\ve x\sim q_h} (K(\ve x), \ell(\ve x, w)) \qedhere
        \end{align*}
    \end{proof}
\end{lemma}

\begin{lemma}[{restating \cref{lemma:psi-susceptibility-expression}}] 
\label{lemma:restated-psi-susceptibility-expression}
    Consider the setting of \cref{sec:granular-theory-introducing-control-parameter}. In 
    particular, let $\phi : \mathcal{W} \to \R$ be a function of the model weights and 
    $\psi : \mathcal{X} \to \R$ a function which satisfies
    \[
        \int _\mathcal{X} q(x) e^{h\psi(x)}\, dx < \infty
    \]
    for all $h$ in some neighbourhood of zero.
    We have
    \[ 
        \chi = -\Cov _{w\sim p_h} \bigl( \phi(w) , \Cov _{ x\sim q}\bigl(\psi( x), \ell( x, w)\bigr) \bigr) + O(\tfrac{1}{n})
    \]
    \begin{proof}
        Using \cref{eqn:general-susceptibility-as-covariance} we have 
        \[
            \chi(h) = -\Cov _{w\sim p_h} (\phi(w), \frac{\partial L ^h}{\partial h}(w)) + O(\tfrac{1}{n})
        \]
        where 
        \[
            L^h(w) = \int _\mathcal{X} \ell(x, w) q_h(x)\, dx
        \]
        and 
        \[
            q_h(x) = \frac{1}{Z_h} q(x)e^{h\psi(x)}, \qquad Z_h = \int_\mathcal{X} q(x)e^{h\psi(x)}\, dx .
        \]
        Note that $Z_{h=0} = 1$ and $\frac{\partial Z_h}{\partial h}  |_{h=0} = \E_{x \sim q}[\psi(x)]$. By the product rule 
        \[
            \frac{\partial}{\partial h} q_h(x)\Big|_{h=0} = \psi(x) q(x) - q(x) \E_{x\sim q} [\psi(x)] . 
        \]
        Then 
        \begin{align*}
            \frac{\partial L^h}{\partial h} \Big | _{h=0} &= \int _\mathcal{X} \ell(x, w) \frac{\partial}{\partial h} q_h(x)\Big|_{h=0}\, dx \\
            &= \E_{x\sim q} [\ell(x, w)\psi(x)] - \E_{x\sim q} [\ell(x, w)] \E_{x\sim q}[\psi(x)] \\
            &= \Cov _{x\sim q}\bigl(\ell(x, w), \psi(x)\bigr).  \qedhere
        \end{align*}
    \end{proof}
\end{lemma}

\clearpage
\section{Experiment details} \label{apx:method-details}

\subsection{Percolation models} \label{apx:percolation-method-details}

\begin{figure}[t]
  \centering
  \includegraphics[width=\linewidth]{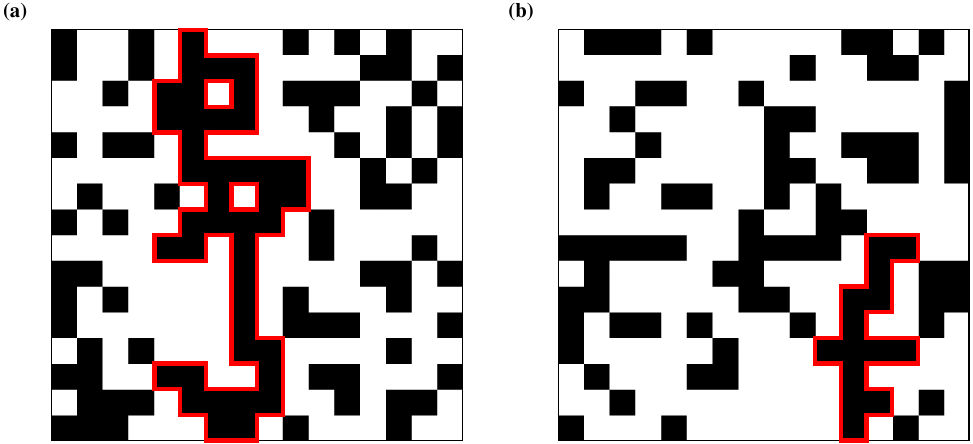}
  \caption{
        An example of percolating and non-percolating states. Each lattice site
        is displayed as a square on a $16\times 16$ grid. Occupied states are coloured 
        black and unoccupied sites are coloured white. \textbf{(a)} A percolating 
        state; the type of states which we train the model to predict. The percolating cluster 
        --- a cluster spanning from the bottom to the top of the lattice --- is highlighted 
        with a red border. \textbf{(b)} A non-percolating state; we do not include such 
        states in the training distribution. In contrast to \textbf{(a)}, there is no 
        connected component of occupied sites spanning from the bottom to the top of 
        the lattice. When evaluating a transformer trained on distributions of percolating 
        states we consider the \emph{tallest cluster} in the generated configuration
        (see \cref{apx:percolation-evaluating-generated-cluster-height}). The tallest 
        cluster in \textbf{(b)} is highlighted with a red border and has height eight because 
        it spans eight rows. The tallest cluster in \textbf{(a)} is the percolating cluster 
        and it has height 16.
  }
  \label{fig:example-of-percolation}
\end{figure}

In this section we give a detailed methodology for the experiments involving percolation 
models, discussed in \cref{sec:percolation-experiments}. 

\subsubsection{Data generation and identifying critical sites} 
\label{apx:percolation-data-generation-and-critical-site}

We sample from the distribution $q_h(\ve x)$ defined in 
\eqref{eqn:percolation-distribution-defn} as follows. First we generate a 
candidate sample $\ve x '$ in which each site's occupation status is independent 
of the other sites. A site in $\ve x'$ is occupied with probability 
$h$ and unoccupied with probability $1-h$. We then check whether $\ve x'$ percolates 
by finding the largest connected cluster of occupied sites and checking whether it 
intersects with both the top and bottom row of the lattice. This process is repeated 
until $\ve x '$ is found to percolate; $\ve x'$ is then returned as a sample from 
$q_h(\ve x)$. An example of a percolating and non-percolating state is visualised in 
\cref{fig:example-of-percolation}.

\begin{remark} \label{remark:on-critical-threshold}
    In an infinite lattice a state \emph{percolates}
    if there exists a connected component of occupied sites with infinitely many 
    sites. 
    For an infinite two-dimensional (square) lattice (i.e. in the limit $N \to \infty$) it is 
    known that percolation happens with probability zero for $h < h_c$ where 
    $h_c\approx 0.59275$ (the exact value is not known; see \citealt[Table 1]{stauffer1985percolation}). At finite $N$ however, percolation occurs with positive probability 
    for $0 < h < h_c$. This allows us to consider distributions $q_h(\ve x)$
    for $h < h_c$ in our experiments with percolation models.
\end{remark}

Let $\ve x$ be a percolating state. To identify if a given occupied 
site $i$ is a \emph{critical site} (\cref{def:critical-site}) in $\ve x$ we form a 
new state $\ve x'$ in which $x'_i = 0$, $x_j' = x_j$ for $j < i$ and $x_j' = 1$ for $j > i$.
We then check whether $\ve x'$ percolates. Site $i$ is a critical site 
in $\ve x$ if and only if $\ve x'$ percolates, according to \cref{def:critical-site}.

\begin{table}[b]
  \caption{Hyperparameter values for the model architecture and training procedure
  of the percolation models.}
  \label{tab:percolation-training-hyperparameters}
  \centering
  \begin{tabular}{lc}
          \toprule
          Hyperparameter  & Value \\
          \midrule
          Total trainable parameters & 14\,839\,809 \\
          Number of layers & 8 \\
          Residual stream dimension & 256 \\
          Number of heads & 4\\
          MLP width & 1024 \\
          Optimizer & AdamW \\
          Batch size & 64 \\
          Weight decay & 0.01 \\
          Gradient clipping norm & 1 \\
          Peak learning rate & $5 \times 10 ^{-4}$ \\
          Final learning rate & $1 \times 10 ^{-5}$ \\
          Total training steps & $2^{18} = 262\,144$ \\
          \bottomrule
  \end{tabular}
\end{table}

\subsubsection{Training and model details} \label{apx:percolation-training-and-model-details}
We train an autoregressive transformer model on a mixture of data from different 
$q_h(\ve x)$, where the lattice size $H=16$. 
That is, for some distribution $u(h)$ with support equal to 
$[0.35, 0.6]$, the training distribution is defined as
\[
    q(\ve x) = \int _{0.35} ^{0.6} q_h(\ve x) u(h)\, dh .
\]
In most experiments we take $u(h)$ to be uniform on $[0.35, 0.6]$; in experiments demonstrating
the effect of biasing the training distribution to smaller and larger values of $h$ we take 
\[
    u(h) \propto \exp(\alpha h)
\]
for $h \in [0.35, 0.6]$ and $u(h)=0$ otherwise, where $\alpha \neq 0$ is a parameter 
controlling the bias towards either end of the interval $[0.35, 0.6]$. As $\alpha \to 0$
the density $u(h)$ approaches the uniform density on $[0.35, 0.6]$. 

When considering a state $\ve x$ as an input to our model, we treat it as a binary string 
by fixing an ordering on lattice sites. Rows are ordered 
from bottom to top and sites are ordered from left to right within each row. 
Since the model uses a causal mask, when making predictions about site $i$, it 
can see all sites in rows below $i$ and all sites to the left of $i$ and in the same row. 
Each site has four input features: the coordinates of the site (normalised to 
$[0,1]^2$), the occupation status of the site (either $0$ or $1$), and the prior 
probability $h$ of occupation. 

We use a standard transformer architecture with blocks consisting of 
multi-head self attention, a two layer MLP with GeLU activation, and layer normalization 
applied twice: before computing attention and in-between the attention and MLP. 
For a positional embedding, we use a 2D version of RoPE which applies rotations 
based on the row and column of each lattice site. The initial embedding of the input 
features into the residual stream is done by treating the four input features as a 
vector in $\R^4$ and applying a two layer MLP with GeLU activation. For each input site, 
the model predicts a distribution on $\{0, 1\}$; the output logit is obtained by applying a 
linear layer to the final vector in the residual stream. 

The model is trained using binary cross entropy loss and the AdamW optimizer for $262\,144$
steps. Unseen data is used every training step. We use a learning rate schedule, starting at 
$0$ with a linear warmup over 2000 steps to a peak learning rate of $5\times 10^{-4}$ followed 
by cosine decay to a final learning rate of $1\times 10^{-5}$ over the remainder of the 
training steps. Gradient clipping and weight decay are used.

Training and architecture hyperparameters are summarised in 
\cref{tab:percolation-training-hyperparameters}.

\begin{table}[b]
  \caption{Hyperparameter values for susceptibility estimation in percolation models.}
  \label{tab:percolation-susc-estimation-hyperparameters}
  \centering
  \begin{tabular}{lc}
          \toprule
          Hyperparameter  & Value \\
          \midrule
          Sampler step size & $1\times 10^{-7}$ \\
          $n\beta$ & 50 \\
          $\gamma$ & 0.5 \\
          RMSProp stability parameter & $1\times 10^{-3}$\\
          RMSProp exponential decay rate & 0.99 \\
          Sampler batch size & 256 \\
          Total steps & 10240 \\
          Burn-in steps & 9728 \\
          Estimation steps (draws) & 512 \\
          Number of chains & 12 \\
          Estimation dataset size & 4096 \\
          \bottomrule
  \end{tabular}
\end{table}

\subsubsection{Estimating susceptibilities} 
\label{apx:percolation-estimating-susceptibilities-details}

We estimate susceptibilities using the covariance expression 
\eqref{eqn:percolation-susceptibility-nested-covariance} found in 
\cref{lemma:percolation-susceptibility-expression}: 
\[
\chi(h) \approx \frac{-1}{h(1-h)}\Cov_{w\sim p_h}(\phi(w), \Cov_{\ve x\sim q_h}(K(\ve x), \ell(\ve x, w))) .
\]
Recall that 
\[
p_h(w) \propto \exp(-n\beta L^h(w)) \varphi(w)
\]
is the quenched posterior. 

We estimate the covariance over $p_h(w)$ using stochastic gradient-based Markov chain
Monte-Carlo sampling; the sampling procedure and the procedure for selecting
sampling hyperparameters are described in \cref{apx:sampler-tuning}. Hyperparameter
values are given in \cref{tab:percolation-susc-estimation-hyperparameters}.

After a burn-in period, we estimate the inner covariance 
$\Cov _{\ve x\sim q_h}(K(\ve x), \ell(\ve x, w))$ each step. To estimate this covariance, 
we use a fixed `evaluation' dataset, which is drawn from $q_h(x)$ independently of the data 
used by the sampler to compute each step. We run a number of independent chains; each 
chain receives independent data for the sampler, independent data to compute the inner 
covariance, and has independent noise. We estimate the variability in the estimate of 
the susceptibility by computing the standard error over the independent chains. 

\subsubsection{Evaluating percolation performance via generated cluster height}
\label{apx:percolation-evaluating-generated-cluster-height}

Here we explain the methodology for evaluating a model's overall ability to make 
predictions about percolation; these results are presented in 
\cref{fig:percolation-data-biasing-results} for models trained on different distributions. 

We assess the model's ability to generate percolating configurations by sampling 
$16\,384$ elements from the mixture of percolation distributions $q(\ve x)$ defined in 
\cref{eqn:percolation-training-distribution}, where $u(h)$ is taken to be the uniform 
distribution on $[0.35, 0.6]$. Note that the uniform mixture is used regardless of the 
training distribution of the model so that we can compare evaluation metrics.
The top $8$ rows of each data sample are removed (recall the lattice size 
is $16 \times 16$) and the model is used to generate the occupation status for the 
remaining $16 \times 8$ sites, using the first $8$ rows as a prompt. 
We then record the height of the tallest cluster in the 
generated state, and take the mean of this quantity over the dataset. An 
optimal model which always generates percolating configurations will always record 
a maximum cluster height of $16$. An example demonstrating the maximum cluster 
height is given in \cref{fig:example-of-percolation}(b). 

For a baseline we use the same procedure described above, except that the top $8$ 
rows are instead completed by randomly generating the occupation status from an 
iid Bernoulli distribution with probability $h$, where $h$ is the probability parameter 
used to generate the sample. 

\subsubsection{Comparing susceptibilities to other quantities} \label{apx:compare-percolation-baselines}

\begin{figure}[t]
  \centering
  \includegraphics[width=\linewidth]{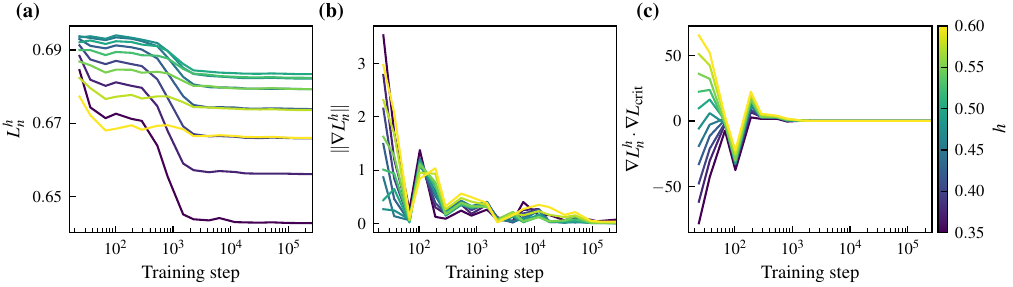}
  \caption{
        Other diagnostic quantities do not predict the same data selection 
        result as cleanly as $\chi_\text{crit.}(h)$; compare to \cref{fig:susc-critical-percolation} (a). 
        \textbf{(a)} The empirical loss $L_n^h$ on iid samples drawn from $q_h$. \textbf{(b)} 
        The norm $\|\grad L_n^h\|$ of the gradient of $L_n^h$. \textbf{(c)} The dot product 
        $\grad L_n ^h \cdot \grad L_\text{crit.}$ between the gradient of $L_n^h$ and the 
        gradient of the loss $L_\text{crit.}$ at critical sites in samples drawn from $q$.
  }
  \label{fig:susc-baselines-percolation}
\end{figure}

To compare to $\chi_\text{crit.}(h)$, we compute other diagnostic quantities over the 
training of percolation models. In \cref{fig:susc-baselines-percolation} we show the empirical 
loss $L^h_n$ on iid samples drawn from $q_h$, the norm $\|\grad L_n^h\|$ of the gradient of this loss, 
and the dot product $\grad L_n ^h \cdot \grad L_\text{crit.}$ between the gradient of $L_n^h$ and the 
gradient of the loss $L_\text{crit.}$ at critical sites in samples drawn from the training distribution $q$. 
These do not predict the same data selection result as cleanly as $\chi_\text{crit.}(h)$. The loss $L^h_n$ 
is non-monotonic in $h$, not suggesting that smaller values of $h$ should be more informative for learning 
percolation. The gradient norm $\|\grad L_n^h\|$ is noisy, with no consistent ordering in $h$ across training 
steps. The gradient dot product $\grad L_n ^h \cdot \grad L_\text{crit.}$ does show similar structure to 
$\chi_\text{crit.}(h)$: the values are consistently ordered by $h$ over training. However, the quantity changes sign 
twice in the period of training where percolation is learned and the different $h$ curves are not as cleanly 
separated as $\chi_\text{crit.}(h)$.   

A more detailed comparison between susceptibilities and other diagnostic quantities is an important direction 
for future work. 

\subsection{Granular material models} \label{apx:granular-method-details}

\subsubsection{Data generation} \label{apx:granular-data-generation}

\begin{figure}[b]
  \centering
  \includegraphics[width=\linewidth]{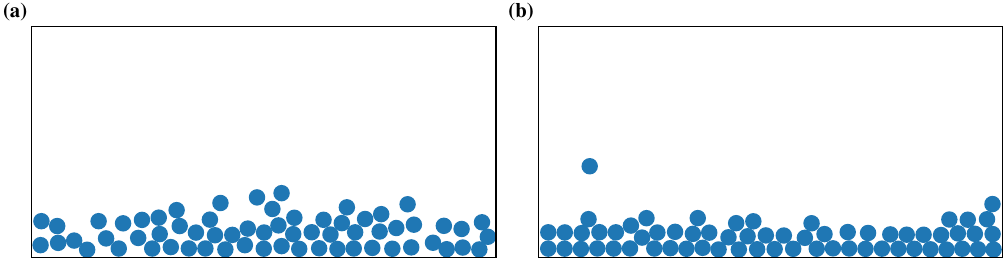}
  \caption{
        Two example states of the granular system: 64 identical spherical particles confined to a
        rectangular box under damped linear spring dynamics. Particle momenta --- which are part of 
        the data of a state --- are not visualised.
  }
  \label{fig:example-of-granular-material}
\end{figure}

We train a model on a distribution $q(x)$ of granular material states defined by the 
following data generation procedure. An initial state is generated by randomly generating 
particle positions in the top half of the box such the particles do not overlap, and 
initial velocities generated according to the standard normal distribution. Damped linear 
spring dynamics are simulated for $5.12\,\mathrm{s}$; see 
\cref{tab:granular-material-parameters} for other quantities defining the granular physics. 
This duration was chosen as it is long 
enough that the system has usually dissipated most of its energy via the damped collisions, 
but short enough that it avoids including a large number of states where the particles are 
static in the dataset. Example states are shown in \cref{fig:example-of-granular-material}.

\begin{table}[b]
  \caption{Hyperparameters of granular material data.}
  \label{tab:granular-material-parameters}
  \centering
  \begin{tabular}{lc}
          \toprule
          Hyperparameter  & Value \\
          \midrule
          Number of particles $N$ & 64 \\
          Particle diameter & 1 \\
          Particle mass & 1 \\
          Collision spring constant $k$ & 1000 \\
          Collision damping coefficient & 10 \\
          Acceleration due to gravity & 9.8 \\
          Box height & 14 \\
          Box width & 28 \\
          Simulated duration & $5.12\,\textrm{s}$ \\
          \bottomrule
  \end{tabular}
\end{table}

\begin{table}[b]
    \caption{Hyperparameters for model architecture and training procedure of 
    granular material models.}
    \label{tab:granular-training-hyperparameters}
    \centering
    \begin{tabular}{lc}
            \toprule
            Hyperparameter & Value\\
            \midrule
            Total trainable parameters & 432\,989\\
            Number of layers & 4\\
            Residual stream dimension & 128 \\
            Number of heads & 8 \\
            MLP width & 128 \\
            Optimizer & Adam \\
            Batch size & 64 \\
            Learning rate & 0.001 \\
            Total training steps & 64000 \\
            \bottomrule
    \end{tabular}
\end{table}

\subsubsection{Training and model details} \label{apx:granular-training-and-model-details}

We train a model with a transformer architecture to predict the future state of the 
granular material after some fixed duration $\Delta t$, given an input state $x$. 
The architecture of the transformer blocks is standard. Each block consists of 
multi-head attention followed by a two-layer MLP with ReLU activation function, and two 
separate layer normalisations applied prior to both the attention and MLP layers. 
Each particle is processed in parallel in its own residual stream, and the attention 
heads are used to transfer information between particles. No attention mask is used. 
We allow the total attention assigned by a head to be less than one by including an 
additional null position with zero-valued key and value vectors.

Boundary information is included in the embedding step. Along each boundary, we define 
fixed points spaced regularly along the boundary. The position vector of each 
point is passed through a learned two-layer MLP (with ReLU activation), where 
the output has dimension equal to the residual stream dimension. 
Each particle's state is passed through a separate two-layer MLP (with ReLU activation), 
and its embedding is augmented by summing the representations of all boundary points 
within a fixed radius.

The predicted state of particle $i$ with position $p_i\in \R^2$ and velocity $v_i \in 
\R^2$ is computed by projecting the output of residual stream $i$ to a vector 
$a_i \in \R^2$ using a linear layer. This is then used to compute 
\[
    v_i' = v_i + \Delta t (a_i + g) \qquad p_i' = p_i + \Delta t v_i'
\]
where $g \in \R^2$ is acceleration due to gravity. This is based on the 
update rule for the semi-implicit Euler integrator, which has good 
numerical stability properties for numerical simulations of damped linear 
spring dynamics 
\citep{samieiAssessmentPotentialsImplicit2013,tuleyOptimalNumericalTime2010}.

The model is trained using mean-squared error loss and the Adam optimizer for 64000
steps. Unseen data is used each training step. We use a constant 
learning rate of $1 \times 10^{-3}$. We do not use either 
gradient clipping or weight decay. 

Training and architecture hyperparameters are summarised in 
\cref{tab:granular-training-hyperparameters}.

\subsubsection{Estimating susceptibilities} \label{apx:granular-susceptibilities-method}

Susceptibilities are estimated in essentially the same way as for the 
percolation models (see \cref{apx:percolation-estimating-susceptibilities-details}), 
using the covariance expression \eqref{eqn:psi-susceptibility-nested-covariance}
found in \cref{lemma:psi-susceptibility-expression}
\[
    \chi = - \Cov_{w\sim p} (\phi(w), \Cov _{x\sim q}(\psi(x), \ell(x, w)))
\]
where $p$ is the (quenched) posterior. Expectations with respect to the (quenched) posterior
are computed by sampling as in the percolation models; the sampling procedure and the
procedure for selecting sampling hyperparameters are described in \cref{apx:sampler-tuning}.
Hyperparameter values are given in \cref{tab:granular-susc-estimation-hyperparameters}.
The inner covariance over the data distribution $q$ is estimated using an evaluation
dataset which is independent of the dataset used for sampling.

In the granular materials models we consider the susceptibility of
the \emph{component-wise loss} \citep[Definition 2.4]{bakerStructuralInferenceInterpreting2025} 
$\phi_C(w) = \delta_C(w)L(w)$, where $C$ denotes 
a component of the model and $\delta_C(w)$ is similar to a delta function which 
we describe below. By a `component' of the model we mean a subset of the parameters 
(\emph{not} a subset of parameter space). In experiments we take $C$ to be the set of all 
attention heads; a parameter is in $C$ if it is used in any attention head of the model. 
Formally, $C$ indicates a factorisation of parameter space $\mathcal{W} = C\times V$. 
Writing $w\in \mathcal{W}$ as $w = (c,v)$ where $c \in C$ and $v \in V$, the notation 
$\delta_C(w)$ indicates the restriction of the integral to parameters only in $C$:
\[
    \int _\mathcal{W} \delta _C(w) L(w) p(w)\, dw = \int _C L(c, v)p(c, v)\, dc .
\]
When estimating integrals involving $\delta_C(w)$ using MCMC, only parameters in $C$ are 
updated and the remainder are held fixed. This allows us to investigate the effect of specific 
parameters on posterior expectations. It is related to the weight-refined local learning 
coefficient introduced by \citet{wang2025differentiation}.

\begin{table}[t]
  \caption{Hyperparameter values for susceptibility estimation in granular material models.}
  \label{tab:granular-susc-estimation-hyperparameters}
  \centering
  \begin{tabular}{lc}
          \toprule
          Hyperparameter  & Value \\
          \midrule
          Sampler step size & $1\times 10^{-6}$ \\
          $n\beta$ & 1385 \\
          $\gamma$ & 1 \\
          RMSProp stability parameter & $1\times 10^{-3}$\\
          RMSProp exponential decay rate & 0.99 \\
          Sampler batch size & 128 \\
          Total steps & 2000 \\
          Burn-in steps & 1800 \\
          Estimation steps (draws) & 200 \\
          Number of chains & 16 \\
          Estimation dataset size & 2048 \\
          \bottomrule
  \end{tabular}
\end{table}

\section{Sampler hyperparameter selection} \label{apx:sampler-tuning}

In both the percolation and granular material experiments, estimating susceptibilities 
involves estimating a covariance with respect to a posterior distribution of the form 
\[
    p(w) \propto \exp(-n\beta L(w)) \varphi(w)
\]
where $L(w)$ is the expected loss at a parameter $w$ and $\varphi(w)$ is a prior distribution. 
We estimate this covariance using stochastic gradient-based Markov 
chain Monte-Carlo sampling, using a similar methodology to \citet{lau2024LLC} 
for estimating the local learning coefficient. Following \citet{lau2024LLC}, 
to estimate the susceptibility at parameter vector $w^*$, we localise the posterior 
using a Gaussian prior
\[
    \varphi(w) \propto \exp(-\tfrac{\gamma}{2}\| w - w^* \|^2)
\]
centred at $w^*$, where $\gamma > 0$ is a hyperparameter of the estimation
process used to control locality. Replacing $L(w)$ with an estimate $L_m(w)$ computed
from a batch of $m$ data points, the gradient term used in sampling --- which is
designed to approximate $\grad \log p(w)$ --- is
\[
    G(w) = -\gamma(w - w^*) - n\beta \nabla L_m(w) .
\]
We sample from $p(w)$ using stochastic gradient Langevin dynamics (SGLD) with the
RMSProp preconditioner \citep{li2016preconditioned}. We run several independent \emph{chains}; the estimate is the mean over chains and its 
standard error is computed across chains. 
The sampler step size, $\gamma$ and $n\beta > 0$ are treated as hyperparameters of the estimation procedure.

Sampling hyperparameters must be tuned so that sampling converges and is stable. 
This is primarily evaluated by visual inspection of \emph{loss traces}: the plot 
of the loss versus sampling step. When sampling is working well, the loss trace increases 
above the value at $w^*$ and levels off into random oscillation about a stable mean. 
Hyperparameters are rejected if the loss traces have large spikes, do not level off, or 
have significant autocorrelation, or if independent chains level off at different values. We aim to find hyperparameters which 
produce stable sampling across all checkpoints of the model for which we want to estimate 
the susceptibility. Hyperparameters for the percolation and granular material models are 
tuned separately. 

In practice, hyperparameter selection was done by sweeping over the sampler step size, $n\beta$
and $\gamma$ values (in log-space), and over several model checkpoints. Loss traces were 
inspected and evaluated according to the above criteria, and additional values were swept 
as needed until a single hyperparameter setting which produced healthy sampling at each selected 
training checkpoint was found. The number of steps and sampler batch size were increased as needed if 
no values for the step size, $n\beta$ and $\gamma$ could be found. Once candidate hyperparameter values were 
found, estimation was run at every checkpoint and the loss traces from sampling runs used to estimate the 
susceptibility were checked and evaluated according to the criteria above. 

The robustness of this hyperparameter selection process and the question of how best to tune hyperparameters 
is an important direction for future work.

\end{document}